# Active Deep Kernel Learning of Molecular Properties from Structural Embeddings

Ayana Ghosh*,[1] Maxim Ziatdinov,[2] Sergei V. Kalinin**[2,3]

[1]Computational Sciences and Engineering Division, Oak Ridge National Laboratory, Oak Ridge, TN, 37831, USA
[2]Physical Sciences Division, Pacific Northwest National Lab, Richland, WA 99352, USA
[3]Department of Materials Science and Engineering, University of Knoxville, Knoxville, TN 37996 USA

**Abstract:**

As vast databases of chemical identities become increasingly available, the challenge shifts to how we effectively explore and leverage these resources to study molecular properties. This paper presents an active learning approach for molecular discovery using Deep Kernel Learning (DKL), demonstrated on the QM9 dataset. DKL links structural embeddings directly to properties, creating organized latent spaces that prioritize relevant property information. By iteratively recalculating embedding vectors in alignment with target properties, DKL uncovers concentrated maxima representing key molecular properties and reveals unexplored regions with potential for innovation. This approach underscores DKL's potential in advancing molecular research and discovery.

Email: *ghosha@ornl.gov; **sergei2@utk.edu

1. **Introduction**

In recent years, the field of molecular discovery[1-8] has experienced a revolutionary metamorphosis, driven by significant advancements in deep learning (DL) models. These sophisticated algorithms have not only accelerated the pace of molecular research but also discerned the advent of a new era in comprehending and forecasting molecular properties. Within the realm of molecular discovery, DL demonstrates its proficiency in deciphering intricate relationships between molecular structures and properties.[8-18] This capability empowers researchers to unravel complex mechanisms[19-27] and expedite the discovery of novel compounds.

Examples abound in the successful application of DL to molecular discovery, particularly in drug discovery.[5,6,28-32] Deep learning models play a pivotal role in swiftly identifying potential drug candidates by predicting their efficacy and safety profiles. These models analyze extensive datasets of molecular structures and biological responses, providing valuable insights that streamline the drug development[23,24,31] process. Moreover, DL models prove invaluable in predicting diverse molecular properties, including toxicity, solubility, and bioactivity. By learning from diverse datasets that encompass molecular structures and experimental outcomes, these models make accurate predictions, significantly economizing time and resources in experimental validation.

The versatility and impact of these models extend across various domains such as quantum chemistry, materials science, prediction of protein structures, and chemical reactions. They moderate the need for computationally expensive quantum mechanical simulations and trial-and-error synthetic chemical routes for efficient exploration of complex molecular interactions, gaining a deeper understanding of dynamic molecular processes. This efficiency is especially beneficial for steering progress in both theoretical chemistry and planning of synthesis.

A typical roster of popular DL models[33-38] includes, but *is not restricted to*, graph neural networks (GNNs),[39-42] recurrent neural networks (RNNs),[22,43-45] convolutional neural networks (CNNs),[46] autoencoders,[47,48] Long Short-Term Memory Networks (LSTMs),[49,50] and attention mechanisms.[51-53] GNNs tend to excel in molecular discovery by representing molecules as graphs, with atoms as nodes and chemical bonds as edges. These networks employ message-passing mechanisms to iteratively update node representations based on their local neighborhoods, enabling them to capture intricate relationships between atoms and predict molecular properties. RNNs are well-suited for sequential data, making them valuable for tasks in molecular discovery where molecular structures can be represented as sequences. CNNs prove effective in molecular discovery when applied to molecular images or grids, using convolutional layers to extract spatial features from molecular structures. Autoencoders contribute to molecular representation learning by encoding molecular structures into a lower-dimensional space and then decoding them back to the original space. This process encourages the model to learn meaningful and compact representations of molecules. LSTMs find utility in predicting molecular behavior over time. Attention mechanisms enhance the interpretability of DL models in molecular discovery by assigning varying degrees of importance to different parts of the input, allowing the model to focus on specific features crucial for the task at hand.

While these models are instrumental in unraveling complex relationships within molecular data, active learning strategies[54] complement them by addressing significant challenges and optimizing the utilization of resources. They boost data efficiency by pinpointing the most informative instances for labeling, thereby enhancing the efficacy of the learning process, particularly when dealing with limited labeled data. This is especially advantageous in contexts where experimental data collection proves to be both costly and time intensive. Active learning mitigates annotation costs by concentrating on instances that yield the most substantial learning improvements, resulting in noteworthy cost savings, especially in domains such as drug discovery and materials science. Additionally, active learning[55-57] adeptly manages imbalanced datasets, navigates diverse regions within the chemical space, and adapts to concept drift over time. By actively selecting demanding instances for annotation, active learning not only contributes to fortifying model robustness but also facilitates transfer learning, amplifying the model's adaptability to related tasks or properties. The iterative characteristic of active learning empowers models to continually enhance their performance, establishing them as invaluable tools for streamlined and effective molecular discovery.

Very importantly, the successes of all the DL models trained as static or within active learning schemes, heavily rely on the molecular embeddings[58-60] such as latent variables in VAEs. These in turn are formed as a compression of static descriptors, for e.g., SMILES (Simplified Molecular Input Line Entry System) and SELFIES (Self-referencing Embedded Strings) to effectively represent and connect molecules to different properties. Active learning models leverage these embeddings to select informative instances for labeling, using the condensed representations to measure uncertainty and guide the labeling process toward the most valuable data points, while enhancing data efficiency, reducing annotation costs, and exploring the chemical space effectively.

However, for molecular discovery, it is crucial to leverage molecular embeddings in a manner that establishes connections with the landscape of molecular properties. Therefore, representing molecules in a low-dimensional latent space linked to specific properties becomes an integral aspect. Autoencoders are often favored for encoding molecular structures into low-dimensional spaces, owing to their past successes. However, the latent representation generated by autoencoders is not inherently linked to any molecular property; rather, it functions as a compressed, abstract portrayal of the input molecular structure. In the autoencoder context, the model is trained to encode a molecular structure into a lower-dimensional latent space and then decode it back to the original structure, with the latent representation intended to capture essential features in a more concise form.

Although the latent representation may encompass information relevant to molecular properties, the autoencoder[61] itself is not explicitly designed to learn or predict specific properties. The typical training objective for an autoencoder is reconstruction, aiming to minimize the difference between the input and the reconstructed output. Consequently, while the latent representation[62,63] may capture structural patterns, these patterns may not directly correlate with explicit molecular properties such as atomization energy, molecular enthalpy, dipole moment.

In this study, we showcase the utilization of deep kernel learning (DKL)[64,65] models with molecular embeddings. Our primary objective is to evaluate the effectiveness of DKL models in predicting molecular properties directly from SELFIES-based one-hot vector representations. We perform this investigation both in standard supervised and active learning[66-68] settings. By using one-hot encodings of SELFIES strings as structural embeddings, we aim to establish a direct mapping between molecular representations and target properties. In addition to property prediction, we analyze the resulting DKL latent spaces to identify how specific molecular properties organize and cluster, enabling the targeted selection or discovery of molecules with desired characteristics.

The model architecture consist of both a deep neural network and a kernel function which can be combined within a Gaussian process (GP).[69-80] These models employ a hybrid architecture, incorporating a deep neural network to extract hierarchical features from input data and a kernel function to capture intricate relationships between these features in a higher-dimensional space. The neural network undergoes training through backpropagation, optimizing its parameters, while the kernel function is simultaneously optimized to measure similarity or distance between points in the transformed feature space. DKL finds applications in regression, classification, and dimensionality reduction, excelling in scenarios involving structured data and non-linear relationships. The synergy of deep and kernel approaches enhances generalization and facilitates effective handling of complex, real-world datasets.

## 2. **Mathematical formulations**

Below is a brief overview of the DKL principles. For a more comprehensive review, we refer the readers to literature[81]. Let's denote input data as $X$, targets as $y$, the parameters of NN as $\Phi$ with output of the NN as $f_{NN}$. Typically, the output is obtained through a series of linear transformations followed by non-linear activation functions. The NN transforms the high-dimensional features space (in this case SELFIES) with a feature mapping.

A standard GP can be defined by its mean function $m(.,.)$ and covariance function (kernel) $k(.,.)$. For input X, therefore, the GP is represented as: $f \sim GP(m(X), k(X, X'))$. Here $f$ is a vector of function values at the points in X and $K(X, X')$ is the covariance matrix. In DKL, $f_{NN}$ becomes a part of the kernel while the mean function remains zero.

$$K_{DKL} = K_{RBF}(f_{NN}(X), f_{NN}(X'))$$

The covariance function $K(X, X')$ measures the similarity of distance between points in the transformed feature space. Commonly used kernels include radial basis function (RBF) kernel with hyperparameter $l$ denoting the length scale of the kernel:

$$k(f_{NN}(X), f_{NN}(X')) = \exp(-\frac{||f_{NN}(X) - f_{NN}(X')||^2}{2l^2})$$

The DKL model parameters (NN weights and RBF kernel length scale) are trained jointly using stochastic variational inference. We then utilize a trained model to derive a posterior predictive distribution for new data points $X_*$:

$$p(y_*|X_*, X, y) = N(\mu_*, \Sigma_*)$$

$\mu_*, \Sigma_*$ are the posterior mean and covariance, respectively, derived from trained model using standard GP formulas.[82]

An example implementation of DKL with GP using an open-access Python-based package, namely GPax 0.1.1 (other package dependencies are also listed in the Supporting Information ) can be found here.

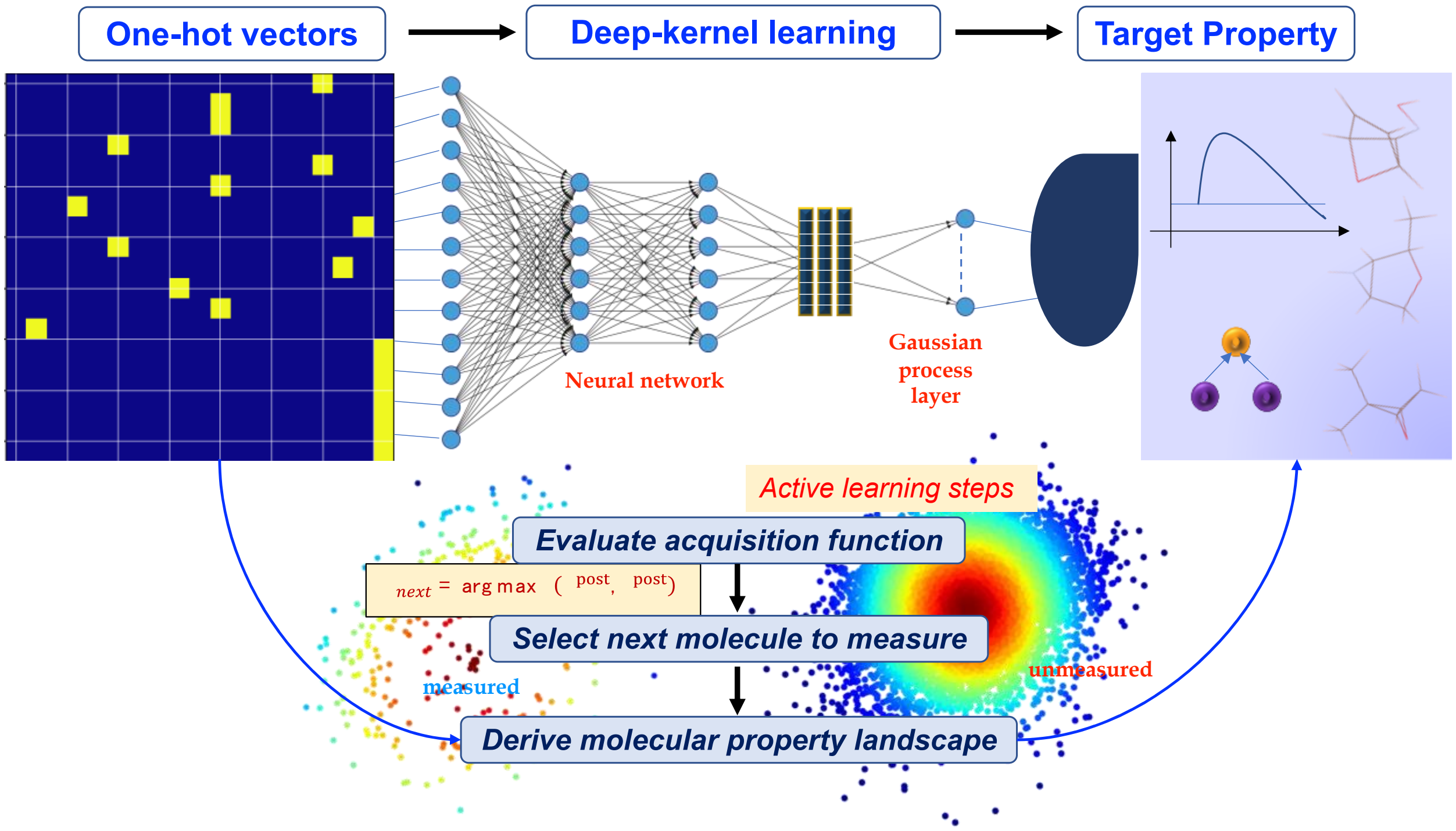

**Figure 1**: Schematic of active deep kernel learning workflow for deriving molecular property from structural embeddings. The first stage includes training of DKL models using hot vectors to predict molecular features. The active learning loop is then initiated with a small subset of points, train models iteratively to predict molecular properties. Each iteration selects the next data point based on an acquisition function, refining the model and comparing predictions.

## 3. Workflow steps

Figure 1 shows a schematic of the workflow as considered in the study. The workflow has been replicated for a randomly selected set of molecules, accompanied by a systematic examination of the resources required to address the scalability aspects of the workflow.

The key steps of the workflow include: (a) training DKL models using one-hot vectors to predict selected molecular features, (b) applying DKL-active learning to predict a target computed via DFT (as available in the QM9 dataset). In each cycle, the model is iteratively retrained, while the complete DKL model from step (a) is continuously updated and refined. Such detailed tracking provides valuable insights into the active learning process, revealing how the chemical space is explored and exploited throughout the learning cycles.

In the active learning segment, a DKL model is iteratively trained on a growing set of molecules represented by SELFIES-based one-hot vectors. At each iteration, the model selects the next molecule by optimizing an acquisition function that balances the predicted property value and associated model uncertainty. This strategy enables targeted exploration of chemical space and progressive refinement of property predictions. We define key hyperparameters, including the number of initial training samples, the number of active learning steps, and batch sizes used for acquisition and prediction. For datasets containing up to 12,000 molecules, we employ an exact Gaussian Process (GP) within the DKL model, ensuring precise covariance estimation and more reliable uncertainty quantification. For the data subsets containing up to 12,000 molecules, we use the exact Gaussian Process (GP) to train the DKL models, ensuring precise computation of covariance matrices. This approach results in highly accurate uncertainty measurements for the trajectories, which directly enhances the effectiveness of molecular discovery.
Example implementations of the DKL models are provided with accompanying notebooks via the Github repository.

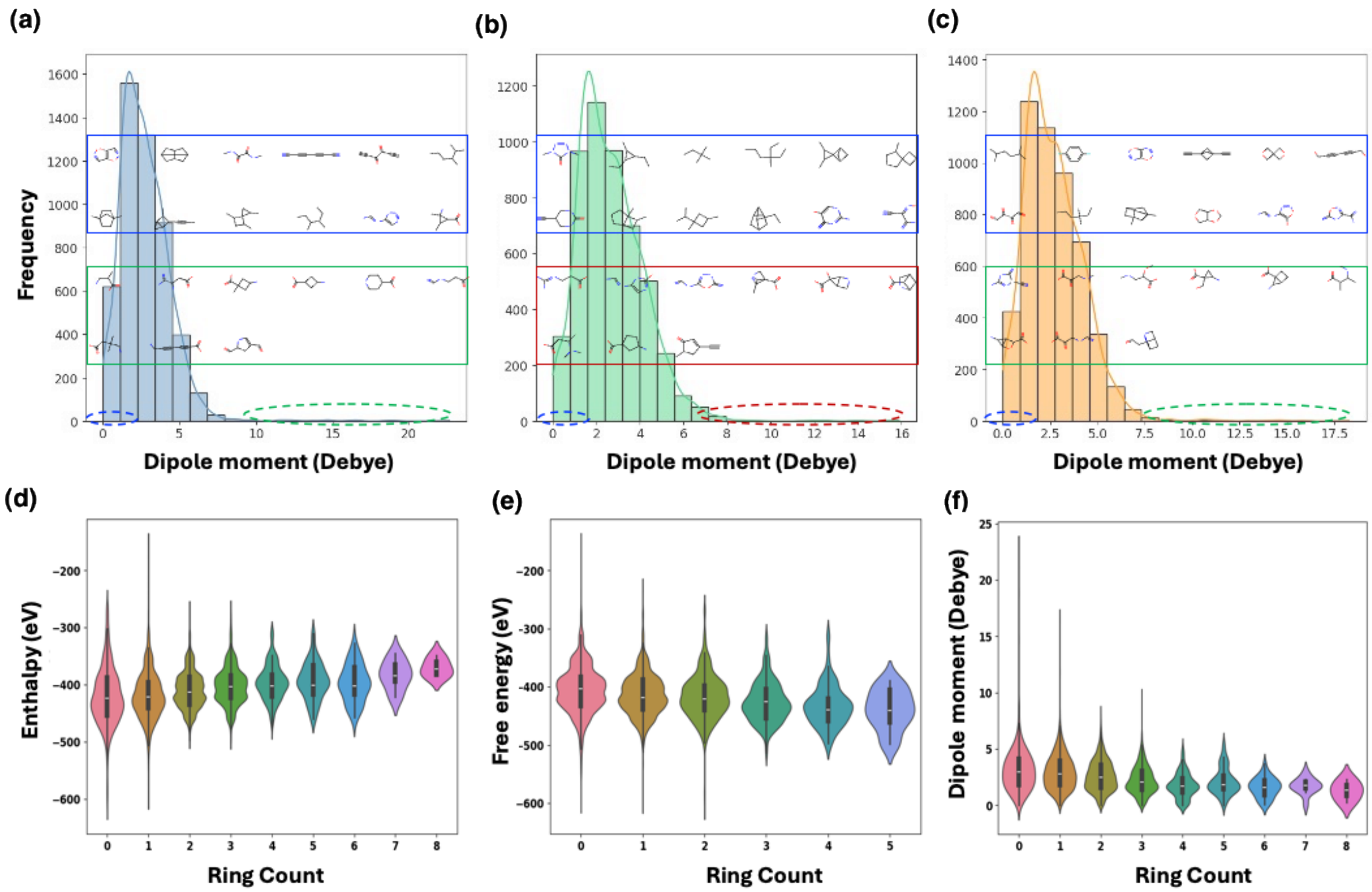

**Figure 2:** The dipole moment (Debye) distribution of molecules is shown for three randomly sampled data subsets of 5,000 molecules (a-c). Molecules from the tails of the distribution—those with negligible dipole moments and relatively high dipole moments—are highlighted in blue and green rectangles, respectively. Panels (d), (e), and (f) display the distributions of selected properties (e.g., enthalpy, free energy, and dipole moment) from the QM9 dataset, plotted against molecular features such as ring counts and rotatable bond counts.

### 4. Dataset & representation

In this study, we have utilized a widely known dataset, QM9 [83], which has served as a benchmark for the development and evaluation of machine learning (ML) models applied to molecular properties prediction. QM9 includes the molecular structures of over 130,000 organic molecules, each represented by its atomic composition and three-dimensional coordinates. The dataset covers a diverse range of chemical compounds, encompassing different functional groups and structural motifs. For each molecule, QM9 provides a set of quantum mechanical properties calculated using density functional theory (DFT) and Hartree-Fock methods. These properties include total energy, enthalpy, free energy, electronic spatial extent, zero-point vibrational energy, and more.

Our goal is to learn molecular features and these pre-computed properties directly from the one-hot vectors. An alphabet of unique SELFIES tokens is constructed from the dataset. Each SELFIES string is then encoded into a fixed-length one-hot vector using this alphabet, with padding applied to match the maximum sequence length. The resulting one-hot vectors are flattened and assembled into the final input feature matrix. These one-hot vectors serve as input features for model training.

To illustrate how DKL models may predict molecular features, we first train models using 5,000 molecules randomly selected from the QM9 dataset. The target molecular features include molecular weight (MW), LogP, number of cyclic structures (ringct), number of rotatable bonds (rotb), all computed using the RDKit Python package. To ensure a representative and reproducible sampling of the chemical space, we generated multiple such subsets using different fixed random seeds. As shown in Figure 2, the resulting distributions capture a broad range of molecular features and properties, supporting consistent model behavior across diverse subsets.

The active learning process is adopted to emulate a realistic setting in which molecular structures or representations are fully defined and available, while property labels—such as those obtained from QM9 dataset directly (pre-computed properties)—are revealed sequentially. Here, the goal is to iteratively improve predictions of computationally expensive properties, including enthalpy, dipole moment, and Gibbs free energy. At each step of the active learning loop, the DKL model selects the most informative molecule, enabling efficient exploration of the chemical space and targeted refinement of model predictions.

## 5. **Results & Discussion**

***Target behavior across dataset***

We have selected 5,000 and then 12,000 molecules randomly from the QM9 dataset for our computations. The target molecular properties chosen are dipole moment, enthalpy, and Gibbs free energy. To implement the DKL workflow, we created three randomized subsets of the same size for each property.

While predominantly featuring weak dipole moments in all the subsets as shown in Figure 2, certain molecules in the first subset (Figure 2a) display notably high dipole moments, indicating enhanced polarity and a propensity for robust interactions with electric fields or polar surroundings. For example, the molecule represented by the SMILES string NC=[NH+]C1=CN=N[N-]1 (~13.73 D) exhibits a substantial dipole moment, primarily attributed to the presence of charged nitrogen atoms. Likewise, CC1C(C([O-])=O)C1(C)[NH3+] has a considerable dipole moment (~14.62 D), likely influenced by the charged nitrogen and oxygen atoms within its structure. Additionally, compounds like [NH3+]C1CC(C1)C([O-])=O and [O-]C(=O)CCNC=[NH2+] showcase elevated dipole moments (~16.54 D and 18.69 D), suggesting the active participation of charged functional groups.

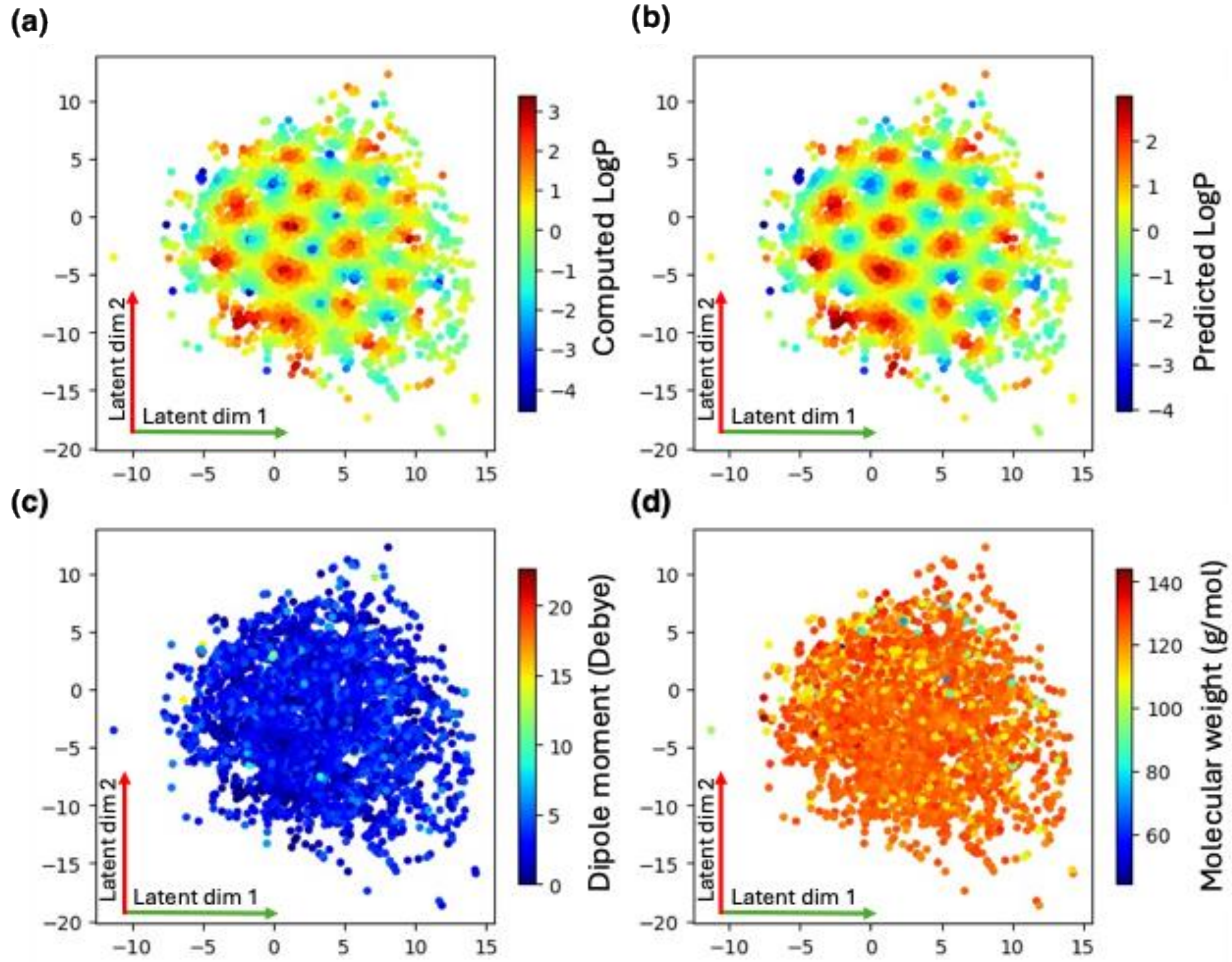

**Figure 4**: Data projected onto the latent space using DKL model (5,000 molecules) trained with one-hot vectors with target feature as LogP. Color maps represent the variation in the feature values across latent space for (a) computed LogP, (b) predicted LogP, (c) dipole moment (D) and (d) molecular weight (g/mol), respectively.

These molecules characterized by heightened dipole moments demonstrate a combination of charge separation, polar bonds, and the presence of charged species, underscoring their potential reactivity and responsiveness to electrostatic interactions. Similarly in the second subset (Figure 2b) N=C1[N-]N=C(NC=[NH2+])O1, for example, features a high dipole moment (~11.89 D) attributed to the presence of charged nitrogen and oxygen atoms. Similarly, compounds like CC1[NH+]2CC1(C2)C([O-])=O and OC1(C2C[NH2+]C12)C([O-])=O showcase high dipole moments (~12.22 D, 12.61 D), likely influenced by charged nitrogen, oxygen, and carbon atoms within their structures. In subset 3 (Figure 2c), there are molecules such as C[NH2+]CC(=O)C([O-])=O and C[NH2+]CC(OC)C([O-])=O exhibit high dipole moments (~12.92 D, 13.82 D), indicating influence of charged functional groups and the presence of polar bonds.

Additionally, consideration of Gibbs free energy (Figure 2(d)), enthalpy (Figure 2(e)), and rotatable bonds (rotb) (Figure 2(f)) provides insights into the thermodynamic and structural aspects of these molecules, enriching our understanding of their reactivity and behavior in various environments. Computational determination of enthalpy and Gibbs free energy for molecules is intricate and computationally demanding. Methods like DFT and coupled cluster, while providing accurate results, require substantial computational resources, especially for complex molecules. Optimization of molecular geometry and calculation of thermodynamic properties involve solving intricate equations, contributing to computational expense. Consideration of temperature and pressure adds further complexity. This has provided us with further motivations to apply the DKL algorithm to predict these molecular properties as additional targets.

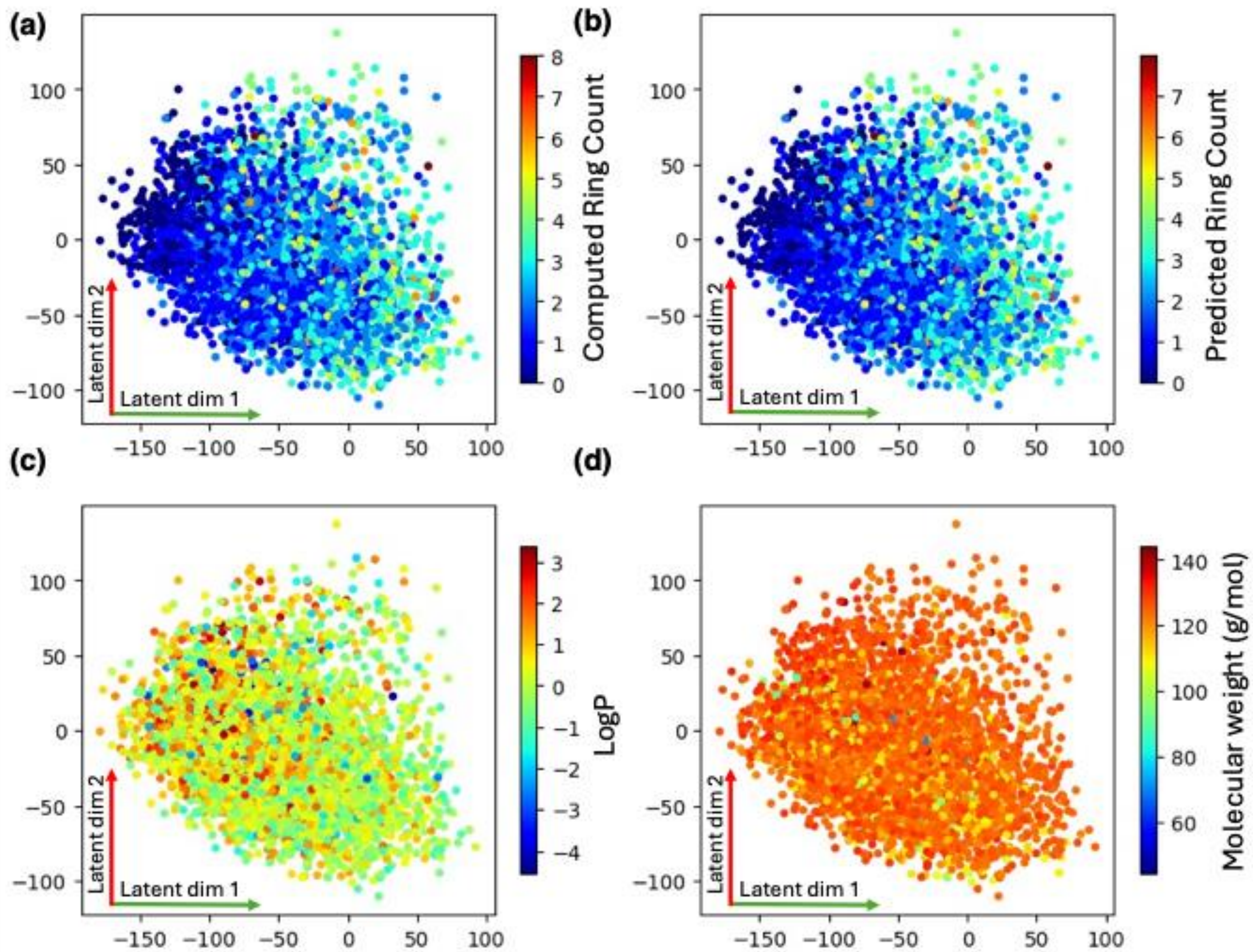

**Figure 5**: Data projected onto the latent space using DKL model (5,000 molecules) trained with one-hot vectors with target feature as total number of rings. Color maps represent the variation in the feature values across latent space for (a) computed ring count, (b) predicted ring count, (c) LogP and (d) molecular weight (g/mol), respectively.

***Standard DKL models***

The standard DKL models are trained in a supervised regression setting using the molecular features such as logP, ring counts values as the primary target variables. Input features are constructed by flattening the one-hot encoded representations of SELFIES strings. While the model is optimized solely with respect to the feature targets, additional molecular features and properties—including dipole moment, molecular weight, and ring count—are preserved for downstream analysis. Following model convergence, we extracted the two-dimensional latent embeddings from the final layer of the deep kernel network. By projecting auxiliary molecular properties onto this latent space, we examine the spatial organization of molecules with respect to their physicochemical characteristics.

As illustrated in Figures 4 and 5, the learned embeddings reveal smooth and localized variations in multiple target properties, suggesting that the DKL model captures underlying structure–property relationships in the molecular dataset. This suggests that the model learns a task-driven molecular representation that generalizes beyond the supervised target, capturing broader structure–property relationships. Moreover, the latent space enables neighborhood-based exploration, allowing for interpretability in terms of molecular similarity and localized property variation.

The standard DKL models demonstrate robust performance across a range of features, properties and data subsets. Predictive mean and variance were computed by averaging the outputs of the Gaussian Process head across mini-batches of latent features, using a batched inference routine that avoids memory bottlenecks. For log P, the model achieves an average MAE of 0.10, RMSE of 0.14, and $R^2$ of 0.98, with predictive variance estimated at 0.04, indicating accurate and confident predictions. Ring count predictions yield an average MAE of 0.16, RMSE of 0.43, and $R^2$ of 0.87, capturing the underlying discrete structure-property relationship with low variance (~0.03). For molecular weight, the model achieves an average MAE of 0.99, RMSE of 1.45, and $R^2$ of 0.96, with a predicted variance of 2.54, while dipole moment predictions produce an average MAE of 0.25, RMSE of 0.32, and $R^2$ of 0.96, with a corresponding variance of 0.12. These results highlight the flexibility of the DKL models trained with the one-hot vector representations in capturing molecular features and properties from learned latent representations.

Figure 6 is an example visual representation of the selected molecules with closest Euclidean distances to molecule represented with SMILE string CC1=C(C)CCCCC1 with the highest LogP value of ~3.286. This down-selection process suggests a correlation between Euclidean distances and predicted LogP. The molecules in the closest neighborhood, i.e., those with shorter Euclidean distances, may be considered more structurally similar. Generally, such a high value of LogP indicated that the molecule maybe somewhat lipophilic which may affect phramokinteic properties. Compounds with favorable lipophilicity characteristics, can be crucial in drug discovery and other applications. Based on the SMILES representations of the molecules in the closest neighborhood, there are a few structural similarities that are evident. All the systems include cycloalkane or cycloalkene structures with varying substituents. They exhibit similar arrangements of carbon atoms in cyclic or bridged cyclic forms, suggesting a degree of structural resemblance with the reference structure.

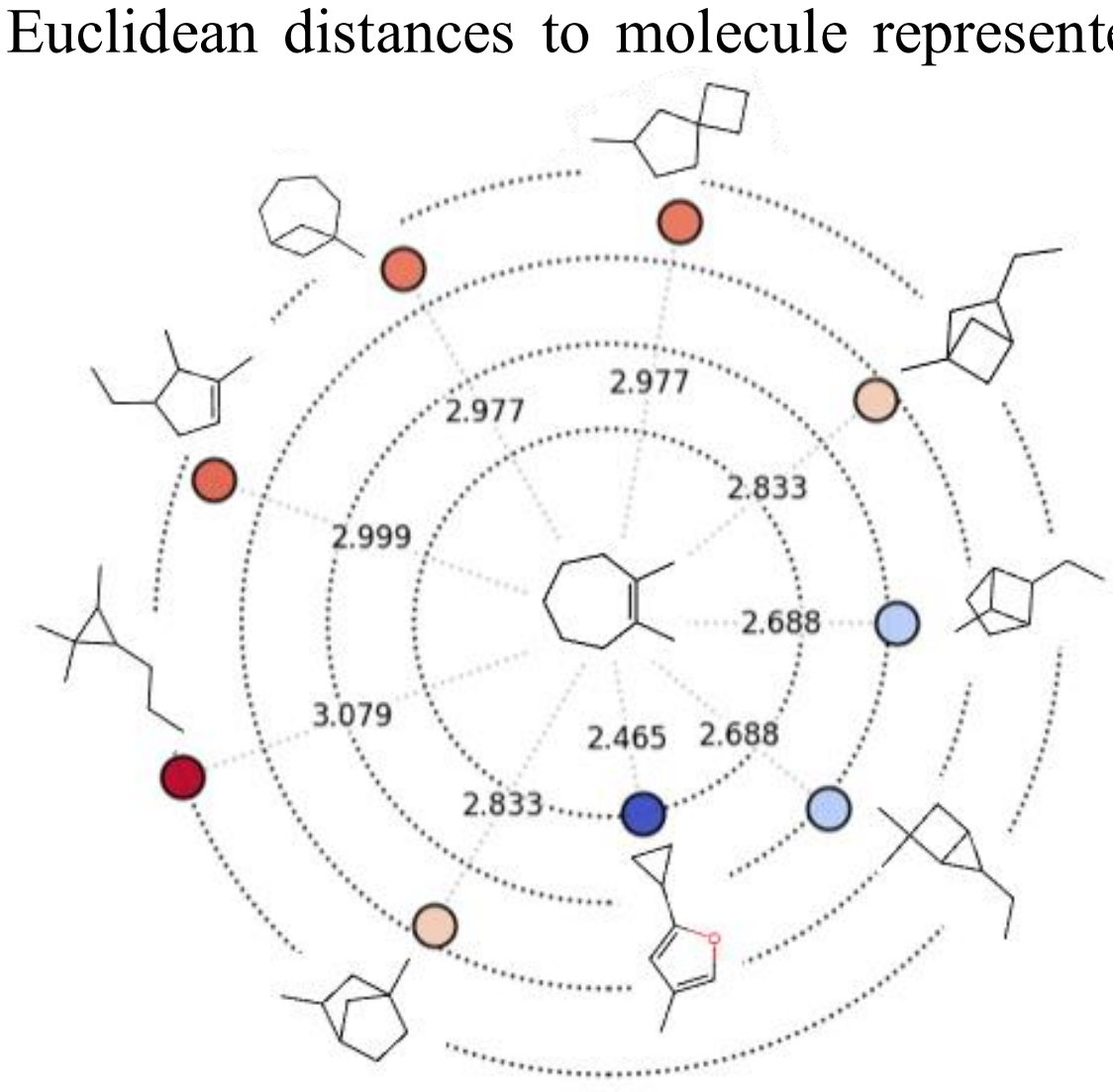

**Figure 6**: Visualization of distinguished neighbor molecules based on predicted log P values by DKL models. The central molecule (star) has the highest predicted log P. Neighbors are positioned by normalized log P distance, with colors indicating log P values and dotted lines showing connectivity.

We have conducted similar studies for other targets such as rincgt and rotb, respectively. For ringct, the down selection leads to curation of molecules with SMILES representations, N#CCC1CCCO1, CC(CC#N)N(C)C, CNC1(COC1)C(C)=O, OC1CC(NC1=N)C#N, CC1C2CC1(C)C1CN21, OC1CC=C2COCC1, CC1CC1(C)CCC#C, CCC1=COC=C1C, COCC1=C(N)C=CO1. They all contain a mixture of functional groups, including nitrogen-containing rings, carbon-carbon double bonds, and cyano groups.

In general, presence of ring like substructures results in increase in molecular complexity, rigidity toward forming conformations, synthesis along with affecting other properties such as its solubility, lipophilicity, and stability. Based on the highest rotb of 6, we have utilized OCCCCOCCO as reference structure to locate the nearest neighbors. The neighborhood comprises molecules such as CCC(CCO)OC=O, OCCOCC(O)C=O, CCC(C)OCCC=O, NCC[NH2+]CCC([O-])=O, CCC(CCOC)C=O, OCC(CO)OCC=O, COCCOCC(C)C, CCCCCCC(C)O, CCCCCOC=NC. They all collectively exhibit a substantial number of rotatable bonds that indicate a certain degree of flexibility to form conformations. They all include multiple aliphatic chains and oxygen-containing functional groups. The presence of oxygen atoms and carbonyl groups suggests potential similarities in their chemical reactivity and interactions.

### ***Active learning with DKL models***

In the next step of the workflow, we perform the active DKL routine for the same set of 5,000 and 12,000 molecules to predict an assortment of DFT-computed targets such as dipole moment, enthalpy and Gibbs free energy as listed in QM9 dataset. Our input representation for molecules remains as the one-hot vectors. The active learning loop is initiated by selecting <2% of the data (100 molecules) from these subsets. Within each iteration, the DKL is trained with the unmeasured points, followed by computation of acquisition function (summation of mean and standard deviation). The optimized point suggested by maximizing (or minimizing) this objective function becomes the next point of measurement.

Additionally, we also retrieve mean and standard deviations of the next point of measurements from one of the standard DKL models trained before within the first step. This is to compare any existing correlations ingrained in the predictions by the standard DKL models endpoints (e.g., LogP, ringct and rotb) with those by the active DKL models endpoints (e.g., dipole moment, enthalpy, and Gibb's free energy). While the standard and active DKL models are trained separately, this cross-referencing strategy provides a conceptual connection between the full and actively sampled models and enables an analysis of shared structural or chemical signals across endpoints. The active learning loop is repeated until all points in the subset have been evaluated, with uncertainty and prediction metrics logged at each step for further analysis. In our current investigation, we have used random sampling for choosing initial 100 molecules to ensure unbiased coverage. Alternative strategies such as diversity- or property-based sampling methods could influence the performance of the model which goes beyond the current scope of our work.

Figure 7 shows the latent space distributions of the measured, unmeasured points with corresponding ground truths as well as prediction uncertainties for predicting enthalpy. The Supporting Information also includes the distributions for the other two target properties. Here, the error is evaluated by taking the difference between predicted from the ground truth (as listed in QM9) and uncertainty representations the respective standard deviations.

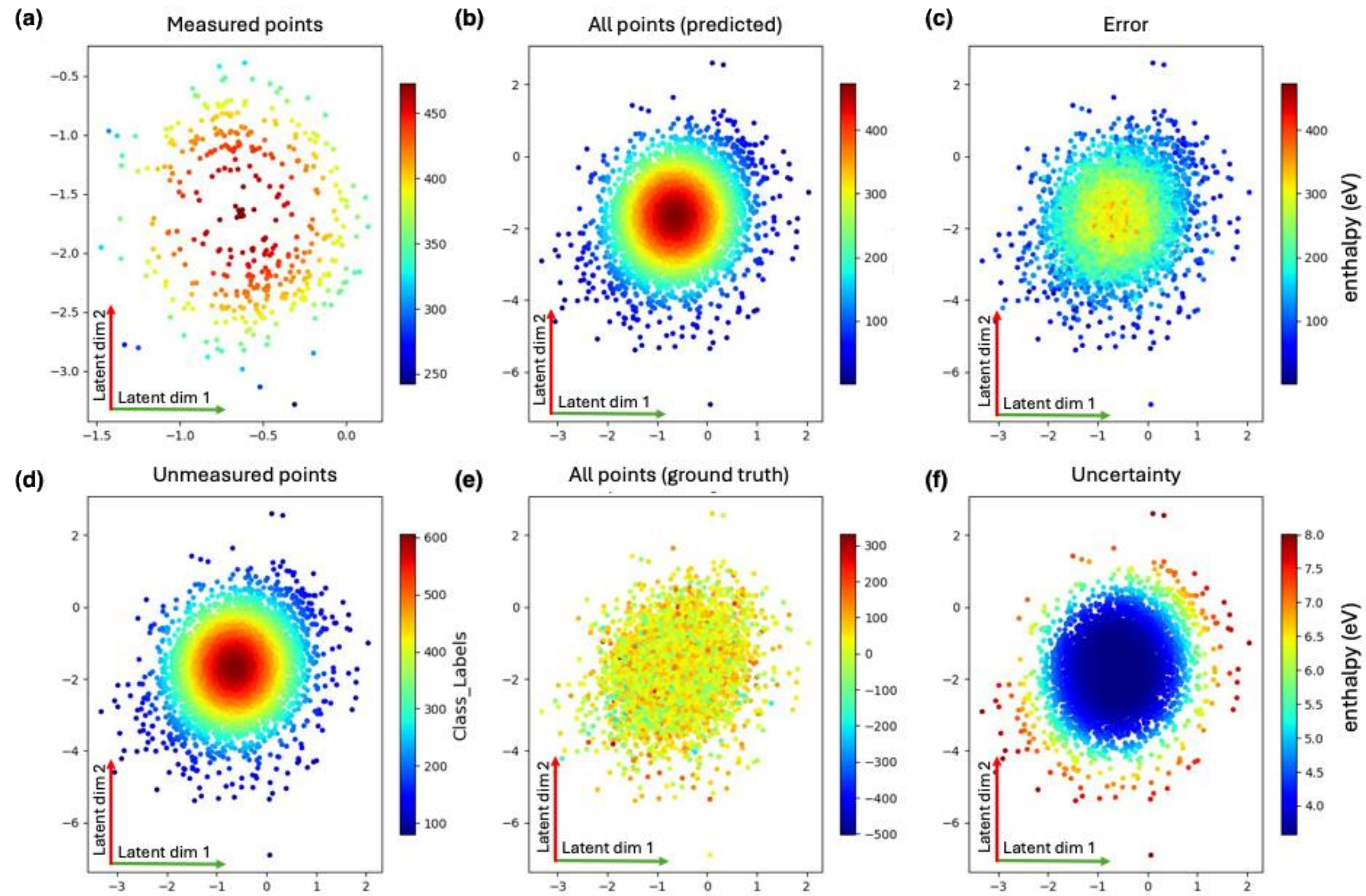

**Figure 7**: Latent space distributions of (a) measured points, (b) corresponding predictions (c) error (predicted - computed), (d) unmeasured points, (e) ground truth and (f) associated prediction uncertainty for enthalpy (eV) of molecules as target using active DKL model.

Distinct regions in the latent space reflect behavior consistent with standard DKL models. While the accuracy appears lower for all targets—as indicated by the uncertainty maps—this is expected due to the active DKL models being trained on a notably small subset of molecules. This limited training data may not fully capture the diverse molecular structures present across the entire dataset. The uncertainty estimates effectively highlight regions where predictions are more challenging, guiding targeted data acquisition. Over the active learning iterations, the model demonstrates significant improvement, balancing accuracy and computational cost by learning meaningful embeddings that support generalization beyond the initially measured samples. To further illustrate this, we include the RMSE reduction over the active learning steps plot in the Supporting Information, highlighting the model's progressive error decrease on the training data.

Lower values of enthalpy and Gibb's free energy typically indicate greater thermodynamic stability of molecules. More specifically, systems with low enthalpy values suggest energetically favorable states within which atoms are already arranged in configurational states with low energy levels whereas low free energy corresponds to greater tendency for the system to move towards a stable equilibrium. On the other hand, most of the molecules within the entire subset exhibit weak polarity, characterized by low dipole moments. This contributes to the challenges in accurately predicting the properties of those with higher dipole moments. The less precise predictions for the latter group arise from the models struggling to effectively capture the embeddings associated with stronger polarity, which is less commonly represented in the overall dataset.

In the supporting information, we have also included the active learning trajectory means with standard deviations as error bars for active DKL models for targets such as dipole moment, enthalpy, and free energy, respectively. The predictions are made in batches of 250 points. This analysis aims to uncover patterns and relationships between different molecular properties, shedding light on how the active learning process influences the exploration of the chemical space and whether certain molecular features are interrelated. Understanding these correlations can enhance our comprehension of the chemical landscape and inform more targeted and effective strategies for molecular discovery. We note that for other datasets with randomly selected molecules, the latent space distributions may vary although the uncertainties in predictions remain within the confidence bound as obtained here.

*Scalability aspect*

The validity of all observations remains consistent for DKL models trained with 12,000 molecules, irrespective of whether they were trained using the complete subsets or within the active learning workflow. We have included these results separately in the Supporting information. We would like to point out that training DKL model with 12,000 data points require substantial use of computational resources. On a single GPU node with 32GB memory, it may take up to an hour to train a standard DKL model. However, there can be potential challenges when computing uncertainty estimates with the exact GP, which necessitates significant memory allocation. We have provided detailed results of a systematic study on computational resources in the Supporting Information. While tasks such as molecular discovery may demand high-precision uncertainty estimates, the active learning workflow can guide subsequent measurements through relative uncertainty estimations.

## 5. **Conclusions**

In summary, we have implemented an active Deep Kernel Learning (DKL) workflow to gain insights into diverse molecular properties. This encompasses features ranging from those readily estimable directly from canonical SMILES string representations to properties requiring computationally expensive first-principles calculations. An essential aspect is the utilization of DKL models, which employ Gaussian process models to establish connections with the target variables. The training of the DKL models has allowed us to derive latent spaces with more distinctive patterns connected to endpoints as compared to those obtained through variational autoencoders or other commonly employed models in the literature.

While molecular features computed from SMILES strings provide initial insights into molecules, relying solely on them may not suffice for accurate predictions of other molecular properties such as enthalpy, free energy, or dipole moment. The one-hot vector representations and the corresponding embeddings created by DKL models prove to be more robust for extracting relationships between molecular structure and properties. Our analyses further demonstrate the potential to derive design principles for targeted properties by uncovering underlying correlations in the data. The dynamically learned molecular embeddings within the active learning framework are designed to address real-time molecular design and the exploration of undiscovered chemical spaces. This has the potential to yield benefits for both the physical sciences and machine learning communities.

## 6. **Acknowledgements**

This research (A.G.) is sponsored by the Laboratory Directed Research and Development Program of Oak Ridge National Laboratory, managed by UT-Battelle, LLC, for the U.S. Department of Energy under contract DE-AC05-00OR22725. This work (S.V.K.) (workflow prototyping) was supported by the US Department of Energy, Office of Science, Office of Basic Energy Sciences, as part of the Energy Frontier Research Centers program: CSSAS—The Center for the Science of Synthesis Across Scales—under Award No.DE-SC0019288, located at University of Washington, DC.

## 7. **Author Contributions**

A.G. implemented the DKL active learning workflow for QM9 dataset and wrote the manuscript draft. S.V.K. implemented the prototype workflow. M.Z. developed the original DKL active learning as implemented in GPax package. All co-authors have participated in manuscript writing and discussion.

## 8. **Code Availability**

The deep kernel learning, active learning workflow illustrative codes can be freely accessed via Github repository.

## 9. **Data Availability**

The datasets used in this work are freely accessible via Github repository.

## 10. **Conflicts of Interest**

The authors declare no competing interests.